\documentclass[acmsmall,authorversion,screen,nonacm]{acmart}

\usepackage{epsfig}
\usepackage{graphicx}
\usepackage{amsmath}
\usepackage{subcaption}
\begin{document}

\title{Hardware-aware mobile building block evaluation for computer vision}

\author{Maxim Bonnaerens}
\email{maxim.bonnaerens@ugent.be}
\orcid{0000-0003-1273-6739}
\affiliation{%
  \institution{IDLab-AIRO, Ghent University - imec}
  \country{Belgium}
}

\author{Matthias Freiberger}
\orcid{0000-0003-2101-6274}
\affiliation{%
  \institution{IDLab-AIRO, Ghent University - imec}
  \country{Belgium}
}

\author{Marian Verhelst}
\orcid{0000-0003-3495-9263}
\affiliation{%
  \institution{MICAS-ESAT, KU Leuven}
  \country{Belgium}
}

\author{Joni Dambre}
\orcid{0000-0002-9373-1210}
\affiliation{%
  \institution{IDLab-AIRO, Ghent University - imec}
  \country{Belgium}
}

\renewcommand{\shortauthors}{Bonnaerens et al.}

\begin{abstract}
In this work we propose a methodology to accurately evaluate and compare the performance of efficient neural network building blocks for computer vision in a hardware-aware manner. Our comparison uses pareto fronts based on randomly sampled networks from a design space to capture the underlying accuracy/complexity trade-offs. We show that our approach allows to match the information obtained by previous comparison paradigms, but provides more insights in the relationship between hardware cost and accuracy.
We use our methodology to analyze different building blocks and evaluate their performance on a range of embedded hardware platforms. This highlights the importance of benchmarking building blocks as a preselection step in the design process of a neural network. We show that choosing the right building block can speed up inference by up to a factor of $2\times$ on specific hardware ML accelerators.
\end{abstract}

%
%

\maketitle

\section{Introduction}
\label{sec:introduction}

In recent years, machine learning models have seen a wide adoption in edge devices. Bringing these models to resource-constrained hardware has accelerated the research in the design of both, computationally efficient neural network models and specialized hardware accelerators that are optimized to execute these neural networks. Many compact neural networks for mobile deployment have been designed over the years, but optimal performance for a given hardware platform can only be achieved when the network is co-designed for it: not every model design choice that aims to improve efficiency does so on every hardware platform. 

However, designing neural networks than can run on a certain target platform with a target latency is a complex task as they have many design choices, ranging from the type of layers used to the spatial resolution and the channel width of each block.
Early work was done by manually designing efficient networks. MobileNetV1 \cite{mobilenetv1} and its introduction of depthwise separable convolutions sparked the beginning of a research field focused on the design of efficient vision models that can be deployed on embedded and mobile platforms. Over the years, improved model architectures were released: MobileNetV2 \cite{mobilenetv2} adds inverted residuals and inverted bottlenecks and MobileNetV3 \cite{mobilenetv3} expands this further by adding the squeeze-and-excitation module \cite{hu2018squeeze} to the bottleneck structure. Another family of efficient models based on group convolutions and a shuffle operator are the ShuffleNets \cite{zhang2018shufflenet, ma2018shufflenetv2}.
All these models use multi-layer {\em building blocks} that are repeated multiple times in the network. Their design process focused mainly on  optimising the layer structure of these building blocks in order to make them hardware efficient. Different variations of a network are typically manually designed \cite{vgg, resnet, ma2018shufflenetv2}, or scaled using a single complexity parameter to trade-off between task performance (e.g., accuracy) and computational cost, e.g., input resolution or channel width \cite{mobilenetv1, ma2018shufflenetv2}. In order to compare such efficient models, most works use curve estimates that show the trade-off between accuracy and a model complexity metric, such as FLOPs or latency, for predefined model variants of a model family. These curves allow practitioners to gain insight in these trade-offs and make the best choices for their design constraints.  

The problem with curve based comparisons is that they only use a handful of model variants, while each model type still has a very large number of hyperparameters. 
Once the initial mobile building blocks had been established, the design process therefore shifted towards neural architecture search to find new state-of-the-art efficient models. Networks optimised with NAS have repeatedly shown to outperform manually designed networks \cite{baker2016designing, liu2018darts, liu2018progressive, real2019regularized, pham2018efficient}. As efficient models are targeted towards mobile and embedded platforms, they are often resource constrained. Hardware-aware NAS incorporates these constraints into the search, which allows finding optimal models for specific target platforms 
\cite{stamoulis2019single, tan2019mnasnet, wu2019fbnet, spos}.
However, the effectiveness of NAS comes with the very large computational cost of training and evaluating many candidate networks. A truly wide search is only feasible when large budgets can be spent on computing resources. 
For this reason, in practice, the search space of current NAS methods is usually constrained, using techniques such as parameter sharing \cite{pham2018efficient}. 
These so called one-shot NAS approaches amortize the training cost of all networks in the search space by training all possible architectures together in a large supernet, reducing the cost of NAS by several orders of magnitude \cite{cai2019once}.
As a result most approaches in NAS now use a search space with a single, fixed building block while searching for other important parameters such as network depth and width or kernel filter size. The choice of the layer structure within the building block itself is mostly based on literature, common practice or previous experience.

As we will show, the relative efficiency of network families that use different building blocks depends, both, on the architecture of the targeted hardware platform and on the desired accuracy range. This means that a true hardware-aware network design approach requires a hardware-aware building block selection method as a front-end before finalizing the full network architecture either through NAS or manual design. To provide a more systematic approach to designing design spaces, \cite{radosavovic2019network} introduce a comparison paradigm based on empirically sampled distribution estimates. While the authors show in their follow up work \cite{radosavovic2020designing} that it can be used to optimize a design space, it does not consider the trade off between complexity within a model family and hardware cost. As we will confirm in this paper, this trade-off offers crucial information, since the best choice can change, depending on the desired accuracy level, or when choosing a more powerful version of a given embedded platform type.


In this work we propose a methodology to evaluate and compare the performance of efficient network building blocks for computer vision in a hardware-aware manner.  As in \cite{radosavovic2020designing} which up until now was the main paradigm to compare design spaces, we use a sampled approach to pre-estimate how well a model family will perform.
However, instead focusing only on the accuracy of a model family, we propose an extension based on a sampled estimate of the pareto front, highlighting the trade-off between complexity and accuracy.  In Section~\ref{sec:edf}, we revisit the work of \cite{radosavovic2020designing}. Our approach is presented in Section \ref{sec:met}, where we also demonstrate why information about the accuracy-complexity trade-off is necessary to make the best choices. We show that our extension allows to match the information obtained by \cite{radosavovic2020designing}, but is better suited to analyze the relationship between hardware cost and accuracy.

Finally, in Section~\ref{sec:evaluation} we use our methodology to analyze and compare some of the most common building block choices on various hardware platforms. Our analysis shows that while certain building blocks constructed with depthwise separable convolutions, inverted bottlenecks and Squeeze-and-Excitation modules may be more efficient FLOPs-wise, they are often not optimal for embedded ML hardware platforms when measuring the actual latency. As such, our approach can be used as a truly hardware-aware efficient building block selection step for the construction of mobile design spaces.

\section{Empirical distribution functions}
\label{sec:edf}
The work of \cite{radosavovic2019network} uses the concepts of design spaces and model families. A \textit{model family} is a collection of related neural network models that share some set of high-level architectural design principles. These can range from very high level principles, such as the convolutional neural networks versus the vision transformer families, to very specific principles such as which specific layers to use and the relation between their depth, width and input resolution in EfficientNets.

A \textit{design space} is a concrete set of architectures that can be instantiated from a model family. A design space consists of two components: a parameterization of a model family, such that specifying a set of model hyperparameters fully defines a network instantiation, and a set of allowable values for each hyperparameter.

To make a robust analysis of a model family over a wide range of network hyperparameters, Radosavovic et al. \cite{radosavovic2020designing} introduced a new methodology: comparing model families using empirical distribution estimates. Instead of comparing handpicked variants of a network, they sample from a \textit{design space} that parameterizes a network family over a wide range of network parameters and train them to collect accuracy metrics, which can be compared  using empirical distribution functions (EDF). The normalized error EDF of a design space with $n$ models with errors $e_i$ is given by:
\begin{equation}
    \hat{F}(e) = \frac{1}{n} \sum_{i=1}^{n} w_i \mathbf{1}[e_i < e]
\end{equation}
$\hat{F}(e)$ gives the fraction of models with error less than $e$, with normalizing factor $w$ to control for model complexity.

While this approach allows to compare entire model families and has shown to be a valuable tool to create better design spaces \cite{radosavovic2020designing}, the metric it uses does not capture differences in accuracy/complexity trade-offs between different model families. This makes EDFs a useful tool when designing model architectures around a certain predefined complexity target.

\section{Capturing trade-offs and search space difficulty} \label{sec:met}
\subsection{Random sampling versus Best models}
The aim of hardware-aware network design is to optimally use the available computing budget to find solutions that are as close as possible to the accuracy/complexity pareto front for the targeted hardware platform. Since, in practice, there is a bound on the total number of network variants that can be evaluated within a reasonable time, it is beneficial to start from a family with similar {\em best} networks but a higher density of {\em good} networks. 

\begin{figure}[h]
\centering
\begin{subfigure}{.32\textwidth}
    \captionsetup{justification=centering}
    \includegraphics[width=\textwidth]{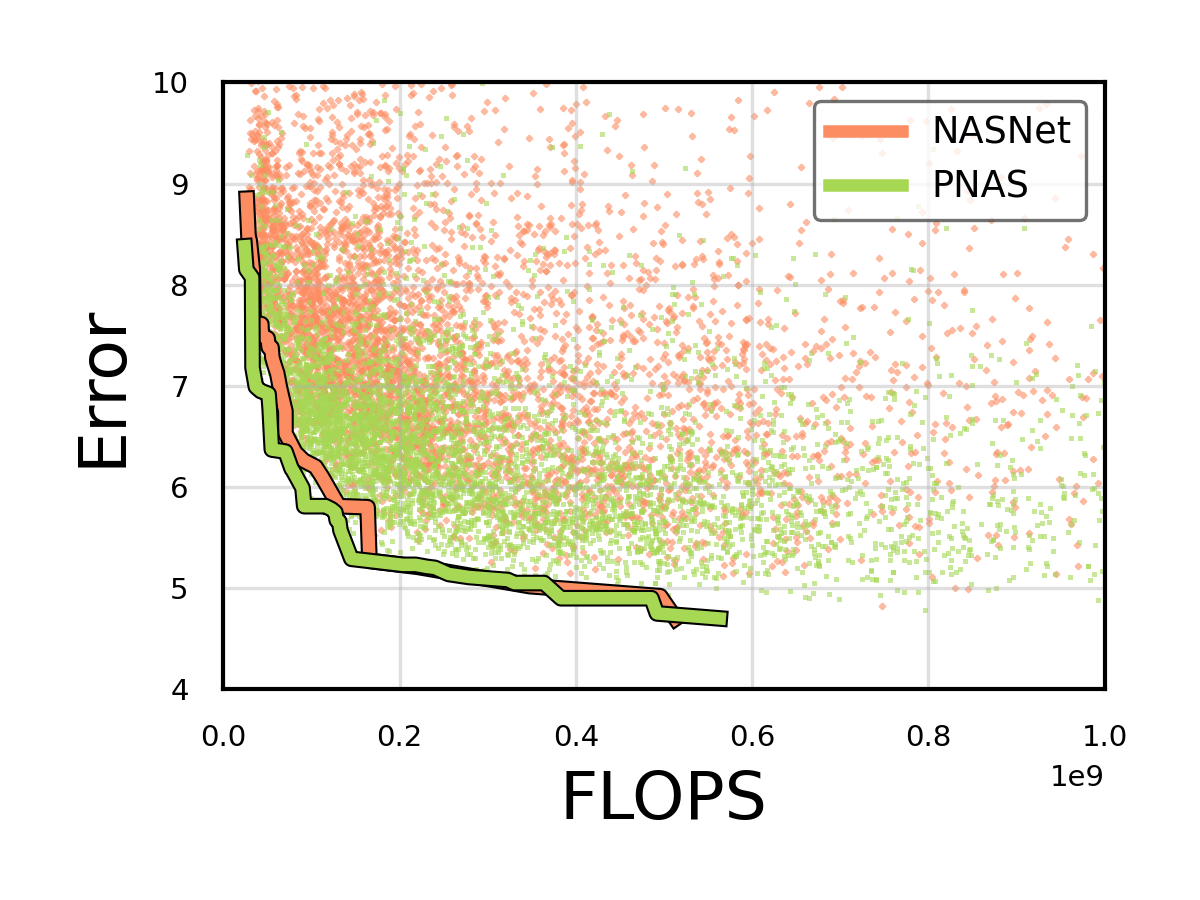}
    \caption{}
    \label{fig:full_distriba}
\end{subfigure}
\hfill
\begin{subfigure}{.32\textwidth}
    \captionsetup{justification=centering}
    \includegraphics[width=\textwidth]{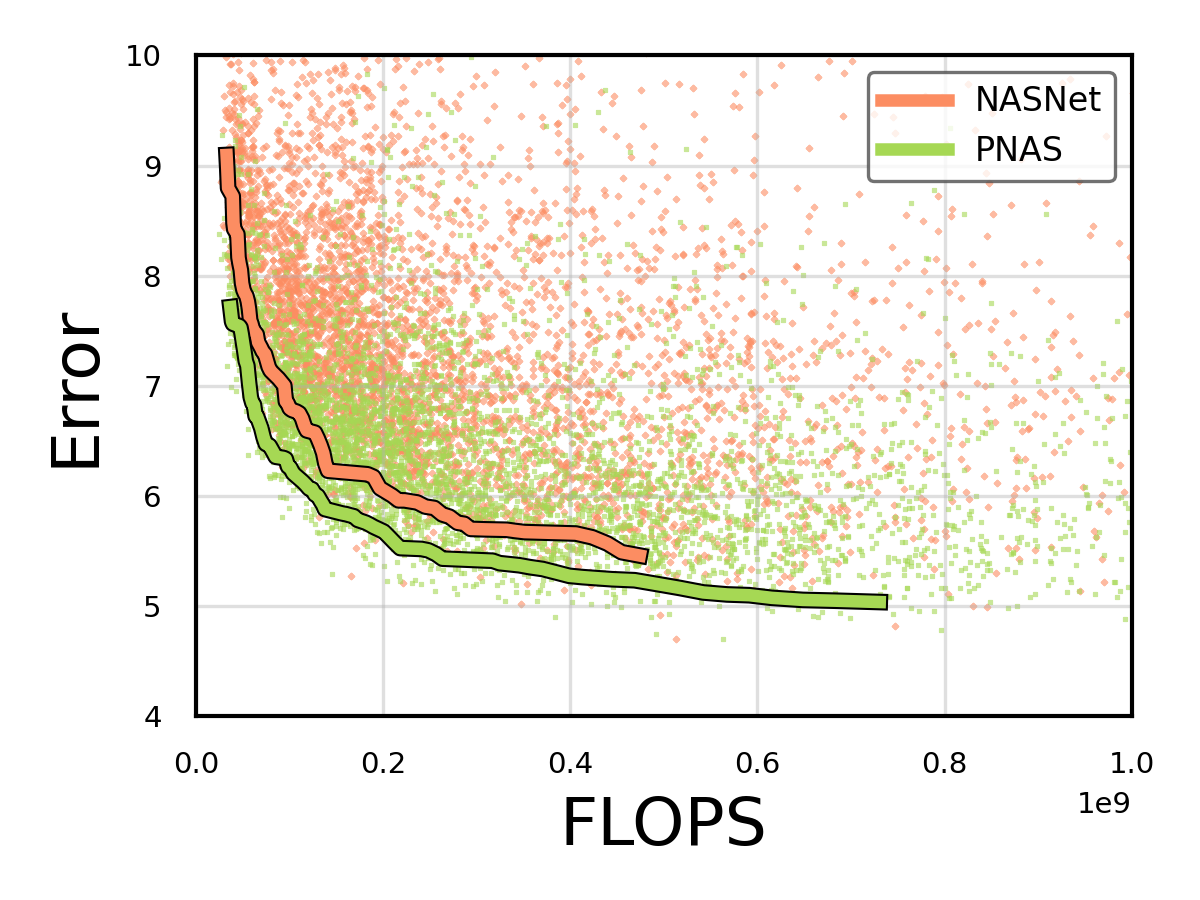}
    \caption{}
    \label{fig:full_distribb}
\end{subfigure}
\hfill
\begin{subfigure}{.32\textwidth}
    \captionsetup{justification=centering}
    \includegraphics[width=\textwidth]{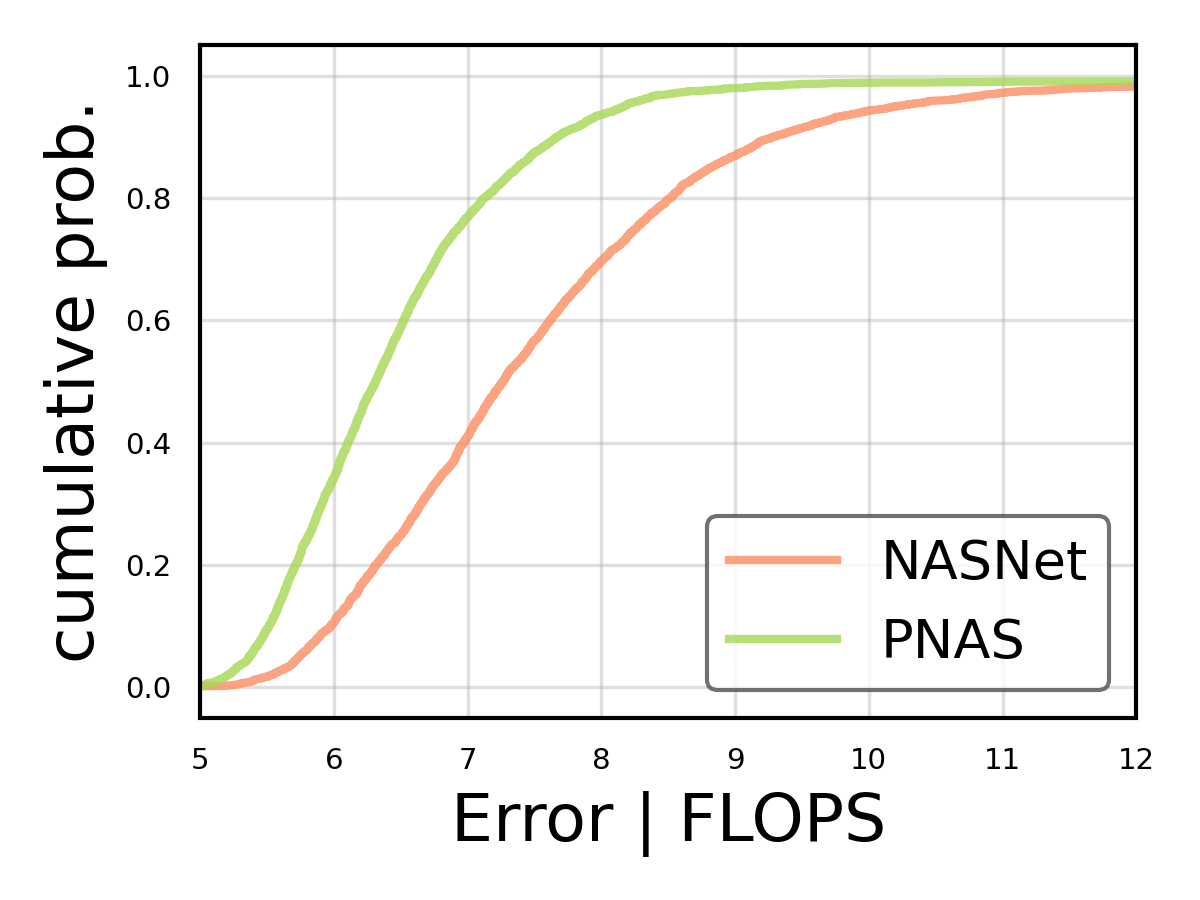}
    \caption{}
    \label{fig:full_distrib_c}
\end{subfigure}
        
\caption{\textbf{(a)} Comparing the best performing models out of 5000 generated ones for the PNAS and NASNet families shows that they can achieve similar accuracies, but in general PNAS will provide good models much faster as can be seen in the normalized error EDF \textbf{(c)}, which is also implicitly captured when creating accuracy/complexity curves (\textbf{b}) through random sampling with only 130 samples.}
\label{fig:full_distrib}
\end{figure}

In \cite{radosavovic2019network}, Radosavovic et al. demonstrated their paradigm of comparing EDFs on different model families by generating the NDS dataset which consists of 5000 models from each corresponding design space. In Figure~\ref{fig:full_distrib} we show  5000 trained models from two of the network families in this dataset, the NASNet \cite{nasnet} and the PNAS \cite{pnas} model families, and highlight the best of those models on an accuracy/complexity curve. 

Comparing only the best models from these larger sampling spaces as depicted in Figure~\ref{fig:full_distriba} indicates that both families are capable of achieving similar results. However when we look at all the models in the scatter plots and the corresponding EDF in Figure \ref{fig:full_distrib_c}, we can see that, when randomly sampled, the PNAS family produces more models with good accuracy than the NASNet family. This means that, for a model designer or NAS-method, is easier to find a well performing model for PNAS than for NASNet. 

Instead of trying to estimate the true pareto front, we propose to use a random sampling accuracy/complexity curve based on a much lower number of samples (only 130). As can be seen in Figure~\ref{fig:full_distribb}, the quality of this approximation decreases more rapidly for the NASNet family. This means that, like the approach in \cite{radosavovic2019network}, it implicitly captures the underlying {\em difficulty} of finding good solutions, but as an added value, this approach maintains the trade-off information. 

In essence our methodology to evaluate model families is done by executing the following steps:  1) create a design space for the model family to be evaluated 2) randomly sample $\pm 130$ models from this design space and train them in a low epoch regime 3) collect hardware costs such as latency from the target hardware platforms 4) create a pareto curve based on the optimal accuracy/complexity points from the sampled models.

\subsection{Sample size}
\label{sec:eval_sample_size}
Training multiple neural networks to evaluate a model family is an expensive step in the process of designing a new model. While it is not necessary to train every sampled model until convergence as a low-epoch regime suffices to gather insights \cite{radosavovic2020designing}, we still aim to minimize the required number of samples.

\begin{figure}[h]
\centering
\includegraphics[width=.8\textwidth]{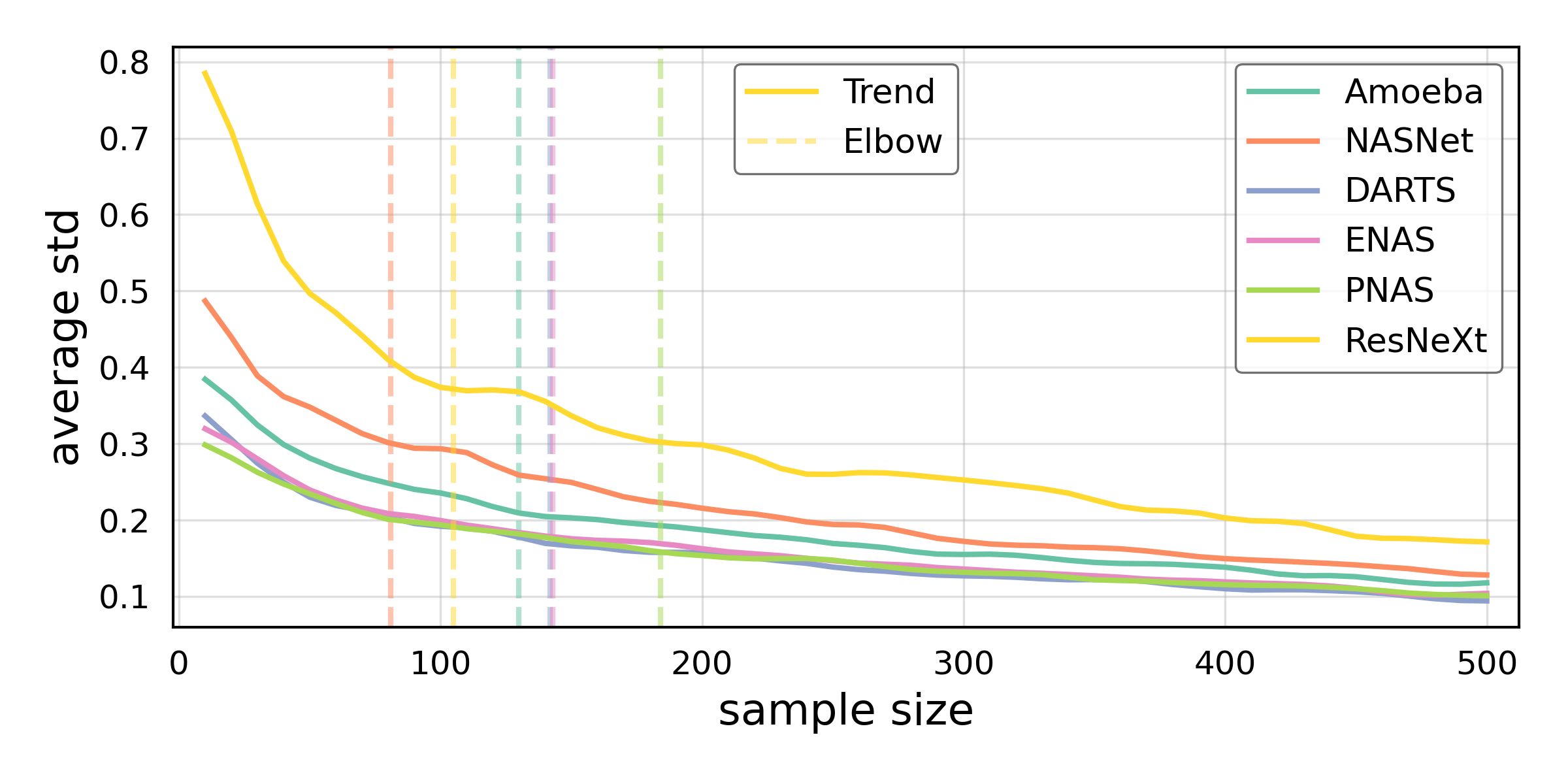}
\caption{\textbf{Sample size.} Trendlines of the average standard deviation for increasing sample size together with their elbow points where the extra cost to train extra networks is not worth it.}
\label{fig:elbows}
\end{figure}

We therefore analyzed the effect of the sampling size on the complexity/accuracy curve. To quantify this, we measured the average standard deviation across the accuracy/complexity curve over a 100 repetitions of a certain sample size. 
To determine the best point in this cost-benefit trade-off we use the Kneedle algorithm \cite{kneedle} to find the elbow point where the cost to train extra networks is no longer worth the expected decrease in noise on the pareto curve. Figure~\ref{fig:elbows} shows the trendlines of the average standard deviation for increasing sample sizes as well as the elbow points. For all families in the NDS dataset this elbow point occurs for a sample size between 80 and 180  with a mean of 128. We therefore recommend a sample size of $\pm 130$ when using random sampling pareto curves.

\subsection{Comparison to empirical distribution functions}
In this section we illustrate why it is beneficial to keep information about the accuracy/complexity trade-off.

\begin{figure}[h]
\centering
\begin{subfigure}{\textwidth}
    \includegraphics[width=\textwidth]{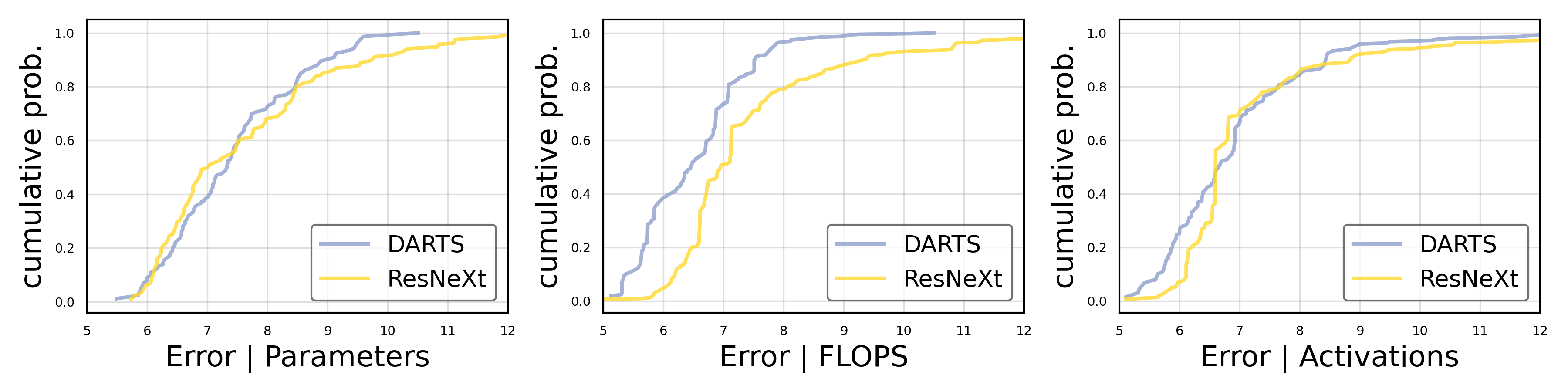}
    \label{fig:edfs}
\end{subfigure}
\hfill
\begin{subfigure}{\textwidth}
    \includegraphics[width=\textwidth]{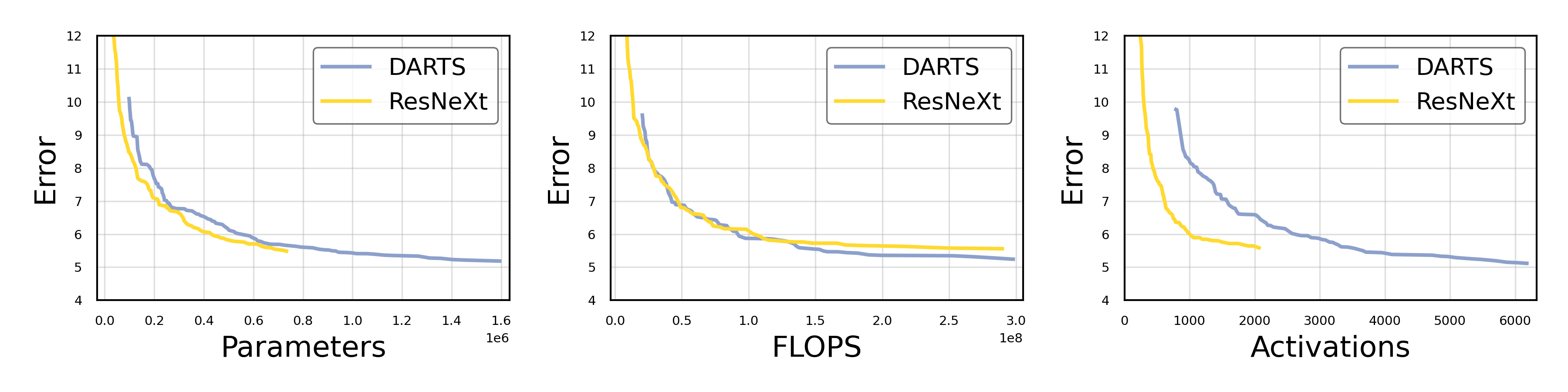}
    \label{fig:paretos}
\end{subfigure}
\caption{\textbf{(top)} The normalized error EDFs highlight that DARTS models produce more highly accurate models. \textbf{(bottom)} Random sampling accuracy/complexity curves show that in the low complexity domain, ResNeXt actually outperforms DARTs.}
\label{fig:edf_vs_r100}
\end{figure}

 We first revisit the results from \cite{radosavovic2019network} for the DARTS \cite{liu2018darts} and ResNeXt \cite{xie2017resnext} families to compare them to our approach. The corresponding normalized EDFs can be seen in Figure~\ref{fig:edf_vs_r100} (top row). To make a fair comparison we only use 130 sampled models for both approaches (\cite{radosavovic2019network} have also shown that any sample size above 100 provides a reasonable estimation). Using their approach, Radosavovic et al. concluded that DARTS and ResNeXt models are similar when normalized by the total number of trainable parameters and that the DARTS family is a better choice than ResNeXt models when normalized by flops  (top left and middle plots). The main insight obtained from these EDFs is that it shows which family provides more higher performing models. However it cannot show how this is related to the complexity of the models. 

In the bottom plots of Figure Figure~\ref{fig:edf_vs_r100}, we show our random sampling accuracy/complexity curves. We can see that the ResNeXt and DARTS families perform similarly in the FLOPs domain until 150M FLOPs, where DARTS models start to outperform ResNeXt models. In the parameter domain we can see that the ResNeXt family produces slightly more efficient models in the low parameter setting but the DARTS family can reduce the error with larger models. Using EDFs becomes less insightful as the difference in the complexity distribution between model families becomes larger. When comparing the same ResNeXt and DARTS models based on their number of activations, which is a better proxy than FLOPs or parameters for certain memory-bound hardware accelerators such as GPUs and TPUs \cite{baker2016designing}, the corresponding EDF still shows that ResNeXt and DARTS models perform similar, while our performance curve shows that the ResNeXt family outperforms DARTS in the lower complexity domain by a significant margin.

\section{Hardware-aware mobile building blocks evaluation}
\label{sec:evaluation}
In this section we show the results of our evaluation of mobile building blocks for convolutional neural networks on various hardware platforms.

\subsection{Mobile convolutional building blocks}
Our focus is on comparing model families that are defined only by their convolutional \textit{building block}, i.e. all models in the family have a structure that consists of a sequence of repeated blocks which can have different structural parameters such as input resolution and output channels but share the same layer structure. The different model families that we compare use the same \textit{design space} and only differ in the used building block.

\begin{figure}[h]
    \centering
    \includegraphics[width=.8\textwidth]{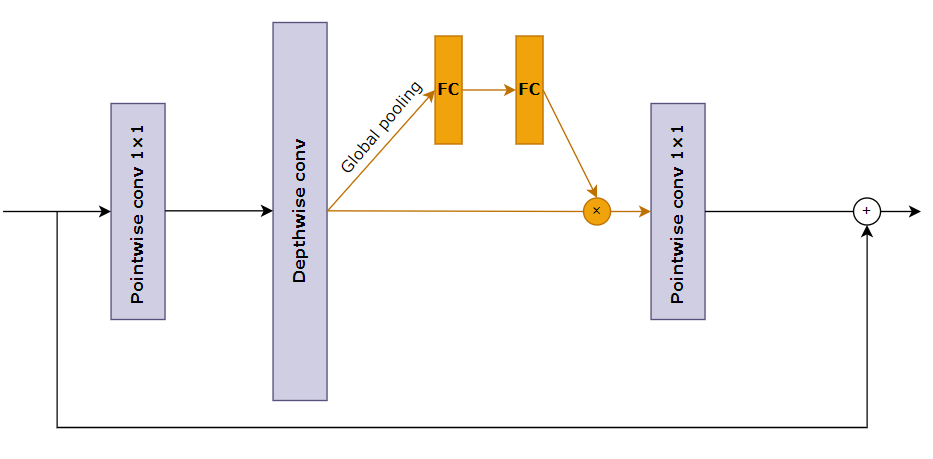}
    \caption{Structure of the MBConv building block, consisting of a depthwise separable convolution with an inverse bottleneck and Squeeze-and-Excitation unit.}
\label{fig:bottleneck_illustration}
\end{figure}

In recent years, one building block has been dominant in hardware-aware designed neural networks \cite{gupta2020accelerator, tan2019mnasnet, mobilenetv3, tan2019efficientnet}: the mobile inverted bottleneck convolution (MBConv). The structure of the building block is illustrated in Figure~\ref{fig:bottleneck_illustration}, it consists of a depthwise separable convolution used in an inverse bottleneck structure and in many cases a Squeeze-and-Excitation unit.

While it has been shown to be a computationally efficient building block, recent work has shown that the large variation within layers of edge models due to the use of various diverse compression techniques results in throughput and energy efficiency shortcomings \cite{boroumand2021google}.

In the remainder of this work we evaluate alternative building blocks which differ in the convolutional layer type, the bottleneck structure and the inclusion of a Squeeze-and-Excitation unit.

\subsection{Setup}
We evaluate building blocks for mobile vision models based on the \textit{RegNet} design space \cite{radosavovic2020designing}. This is defined by 4 parameters: depth $d$, initial width $w_0$, slope $w_a$ and quantization $w_m$. Instead of making the width of every block in the neural network a hyperparameter, the RegNet design space parameterizes the block widths as $u_j = w_0 + w_a \cdot j$ for $ 0 \leq j < d$. This block width is additionally quantized through the hyperparameter $w_m$ such that the network only increases width at each of the 4 stages, where each stage consists of a sequence of identical blocks (see \cite{radosavovic2020designing} for more details).

For all evaluations we sampled 130 models (as described in Section~\ref{sec:eval_sample_size}) from the design spaces created by the evaluated building block, and train each model for 10 epochs. 
The models are trained using SGD and a cosine learning rate schedule with initial learning rate 0.05 a weight decay of $10^{-4}$ and a batch size of 128 on an RTX 3090.

All evaluated networks  are trained on the Visual Wake Words dataset \cite{chowdhery2019visual}. This is specifically designed for vision models in embedded applications. It represents the vision use-case of identifying whether a person is present in an image or not. The dataset is derived from the publicly available COCO dataset, and provides a realistic benchmark for tiny vision models.

\subsection{Hardware platforms}
We evaluated the different mobile building block choices on a range of hardware platforms.
The platforms evaluated are: a CPU and GPU found in modern mobile phones such as the Pixel 4, a dedicated hardware accelerator for edge devices (i.e. Intel Movidius Myriad X Vision Processing Unit (VPU)) and an embedded GPU from NVIDIA found on the Jetson Nano. We also include a benchmark on a server-grade GPU to highlight the difference between choices for mobile deployment versus cloud deployment.
Table~\ref{tab:hw_setup} gives an overview of the hardware platforms and also includes the inference framework used as it also influences the inference speed.

The latencies from the mobile CPU, mobile GPU and VPU are obtained using nn-Meter \cite{nnmeter}, a highly accurate latency prediction library. 

\begin{table}[h]
    \centering
    \begin{tabular}{c|c|c|c}
        \hline
         & Device & Processor & Framework \\ \hline
        mobile CPU & Pixel4 & CortexA76 CPU & TFLite \\ 
        mobile GPU & Pixel4 & Adreno 640 GPU & TFLite \\ 
        VPU & Intel NCS2 & MyriadX VPU & OpenVINO \\ 
        embedded GPU & Jetson Nano & NVIDIA Maxwell GPU & TensorRT \\ 
        server GPU & Server & NVIDIA RTX 3090 GPU & PyTorch \\ \hline
    \end{tabular}
    \captionsetup{justification=centering}
    \caption{Evaluated hardware platforms}
    \label{tab:hw_setup}
\end{table}

\subsection{Depthwise separable convolutions versus standard convolutions}
Since its introduction in MobileNet \cite{mobilenetv1}, depthwise separable convolutions have been the de facto building block for efficient vision models. A depthwise separable convolution factorizes a standard convolution into a depthwise convolution and a $1 \times 1$ convolution called a pointwise convolution. This factorisation reduces the computations by a factor of $\frac{1}{C_{out}} + \frac{1}{K^2}$, where $C_{out}$ is the number of output channels of the convolution and $K$ the kernel size. For the commonly used $3 \times 3$ kernel size this means a reduction by a factor 8 to 9. While depthwise separable convolutions were initially introduced for small vision models, they became standard in vision models across all complexity ranges, e.g. the very large vision model EfficientNet-L2 with 480M parameters and 585B FLOPS also uses them.

Although standard convolutions may be more computationally expensive, they can, in certain cases, utilize the hardware resources better. 
Hardware accelerators for DL commonly use wide single-instruction multiple-data (SIMD) processing units to achieve a high FLOP/S throughput. However high throughput can only be achieved if there is enough data re-use to fully utilize the processing units. In depthwise convolutions the data re-use is much lower than in standard convolutions as each input feature map is only used in the computation of its corresponding output feature map. During training, the resource utilization can be improved by using large batch sizes, obtaining data re-use with the convolutional kernel weights. However, during inference where typically batch sizes of 1 are used, this puts a memory bottleneck on the hardware accelerator.
As a result, fused versions of the MobileNetV2 building block (Fused-MBConv) where the first pointwise and depthwise convolution are fused into a standard convolution have recently been included in some hardware-aware neural architecture searches \cite{xiong2021mobiledets, gupta2020accelerator, tan2021efficientnetv2} and shown to be part of the best models on certain hardware platforms. 

\begin{figure}[h]
    \centering
    \includegraphics[width=\textwidth]{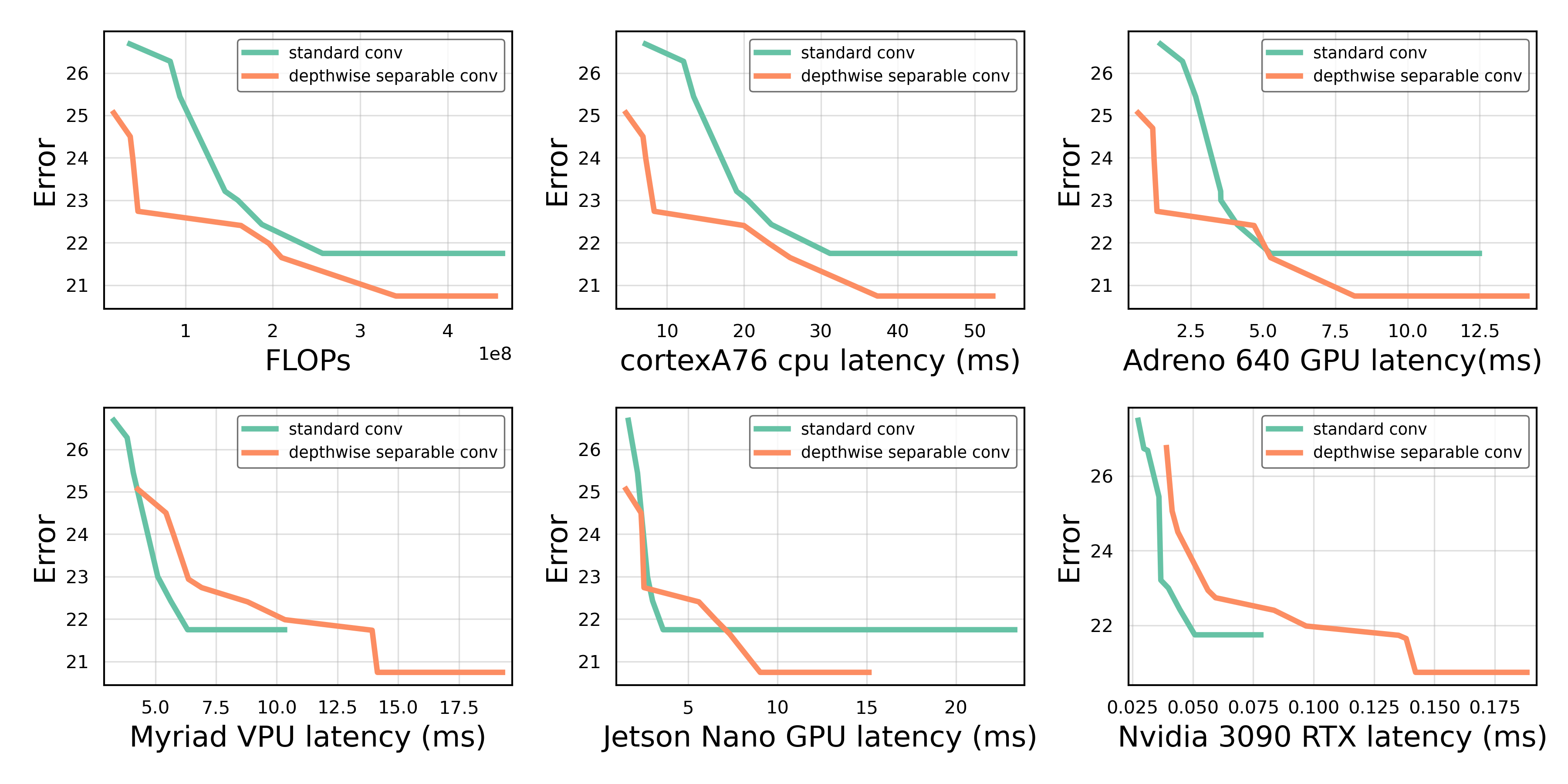}
    \caption{Depthwise separable convolutions are efficient when evaluated by FLOPS (upper left) but this does not translate to faster inference on all embedded hardware platforms.}
\label{fig:dw_vs_conv}
\end{figure}

Figure~\ref{fig:dw_vs_conv} compares the randomly sampled pareto front of the design spaces based of the depthwise separable convolution with the standard convolution. It can be seen that depthwise separable convolutions indeed outperform standard convolutions when comparing FLOPs. The performance curves for the mobile CPU and mobile GPU have similar trend lines and also favour the depthwise separable convolutions. For the Jetson Nano embedded GPU, however, we see that the theoretical advantage of standard convolutions no longer translates into faster latency. Instead, standard convolutions outperform depthwise convolutions as the complexity grows. On the VPU and server-grade GPU the intial order is reversed and standard convolution executes faster than the depthwise separable convolution.

To quantify the difference, we selected models with equal FLOPs from both families and compared the latencies on the different hardware platforms. On the mobile CPU, models from both families with equal FLOPs also have about equal latencies but on the VPU the inference time of the depthwise separable convolution models is $2 \times$ longer than the standard convolution models. A similar trend was seen on the embedded GPU where the latencies of those models were a factor $1.85 \times$ longer.

\subsection{Grouped convolutions}
\begin{figure}[h]
    \centering
    \includegraphics[width=\textwidth]{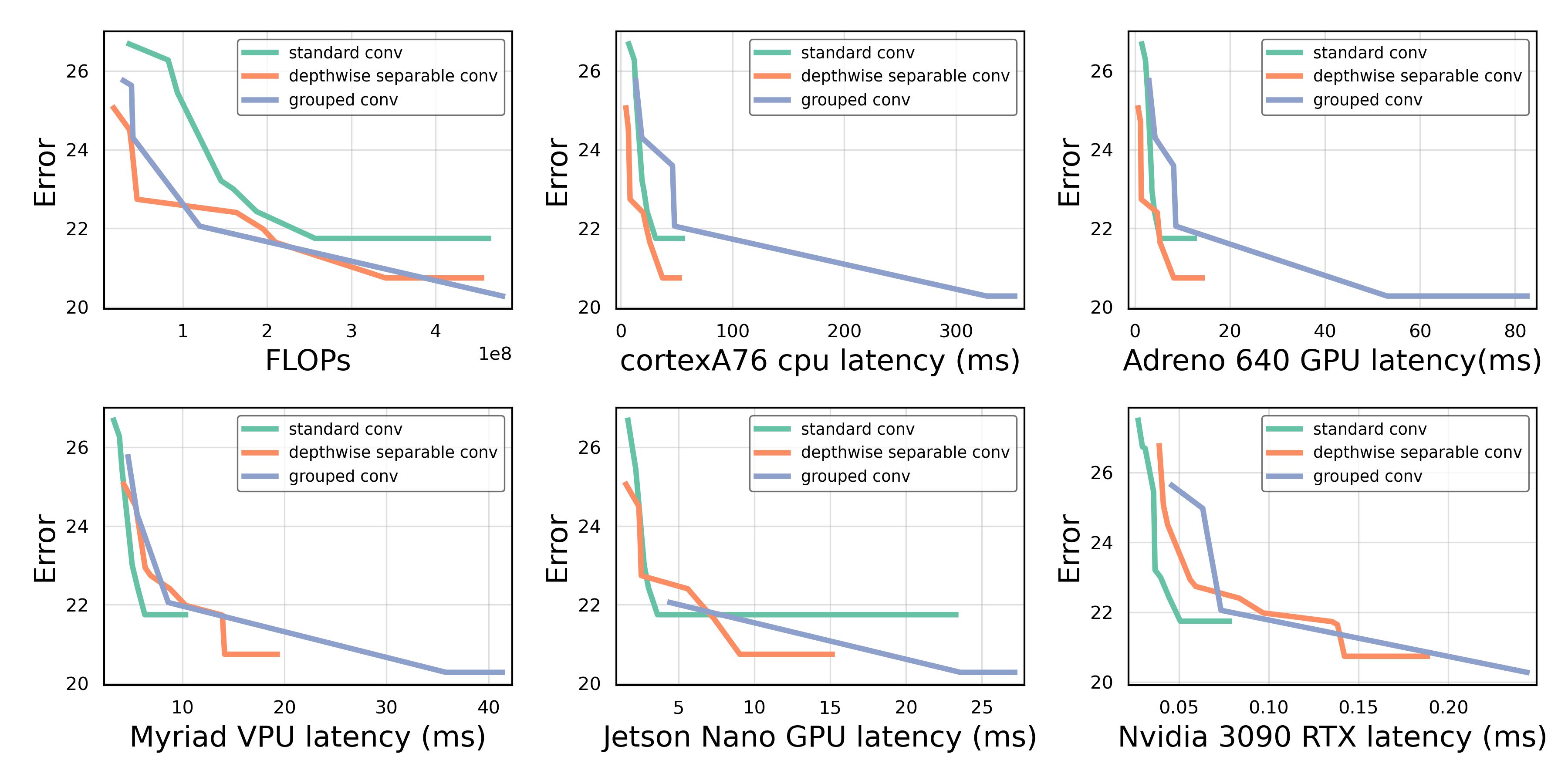}
    \caption{While grouped convolutions should in theory be able to combine advantages from depthwise and standard convolutions, current implementation across various hardware accelerators and inference platforms fail to leverage these advantages making grouped convolutions not an optimal choice.}
    \label{fig:grouped_conv}
\end{figure}

In grouped convolutions the input channels are divided in multiple groups over which a normal convolution is performed, which puts their cost in FLOPS between those of depthwise separable and standard convolutions. This is confirmed in Figure Figure~\ref{fig:grouped_conv} (top left). In fact, depthwise convolutions can be considered as a special case of grouped convolutions, where the number of groups equals the number of input and output channels. Similarly, standard convolutions are the special case where there is only 1 group. 

In theory, grouped convolutions should be able to improve hardware utilization through their improved data reuse while still using significantly fewer FLOPs than standard convolutions. However, current implementations in most deep learning frameworks fail to leverage these advantages, which in practice makes grouped convolutions slower than their standard convolution counter part \cite{gibson2020optimizing}.
Figure~\ref{fig:grouped_conv} shows that, indeed, grouped convolutions are never the most promising solution on any of the tested hardware platforms. This means that, for most target latency ranges and platforms using the much larger search space offered by grouped convolutions will rarely be beneficial. Instead, it is more efficient to narrow down the model design space to either depthwise separable convolutions or standard convolutions, based on their relative performance in trade-off plots like Figure~\ref{fig:grouped_conv}

\subsection{Bottleneck versus inverted bottleneck}
Bottlenecks were introduced together with residual connections in ResNets \cite{resnet}. The main idea behind this is to lower the computational cost of each building block by performing the convolutions at a lower dimension, such that more building blocks can be stacked and deeper networks can be created without significantly increasing the computational cost.

A standard bottleneck structure consists of three convolutional layers: a $1 \times 1$ pointwise convolution to reduce the channel size, a standard convolution to improve the features, and a final $1 \times 1$ pointwise convolution to restore the channel dimensions. A parallel residual path connects the input and output of this block.
In MobileNetV2 \cite{mobilenetv2} the \textit{inverted} bottleneck was introduced where the residual connection is moved to connect the bottlenecks. In practice, this means that an inverted bottleneck consists of a pointwise convolution to expand instead of reduce the channel dimension, a (depthwise) convolution and a final pointwise convolution to restore the original channel dimension. The motivation for this design was that it is more hardware efficient as only the bottleneck lower dimension tensors need to be fully saved to memory as the intermediate higher channel dimension tensors are only used in depthwise convolutions. This means the tensor can be split into smaller ones.

A third possibility is to use no bottleneck, since the RegNet paper found that their best models used a bottleneck expansion/reduction factor of $1.0$ \cite{radosavovic2020designing}.
When using no bottleneck we also drop the final pointwise convolution present in bottleneck structures since it is no longer required to make channel sizes match. All our building blocks in this comparison make use of depthwise convolutions.
\begin{figure}[h]
    \centering
    \includegraphics[width=\textwidth]{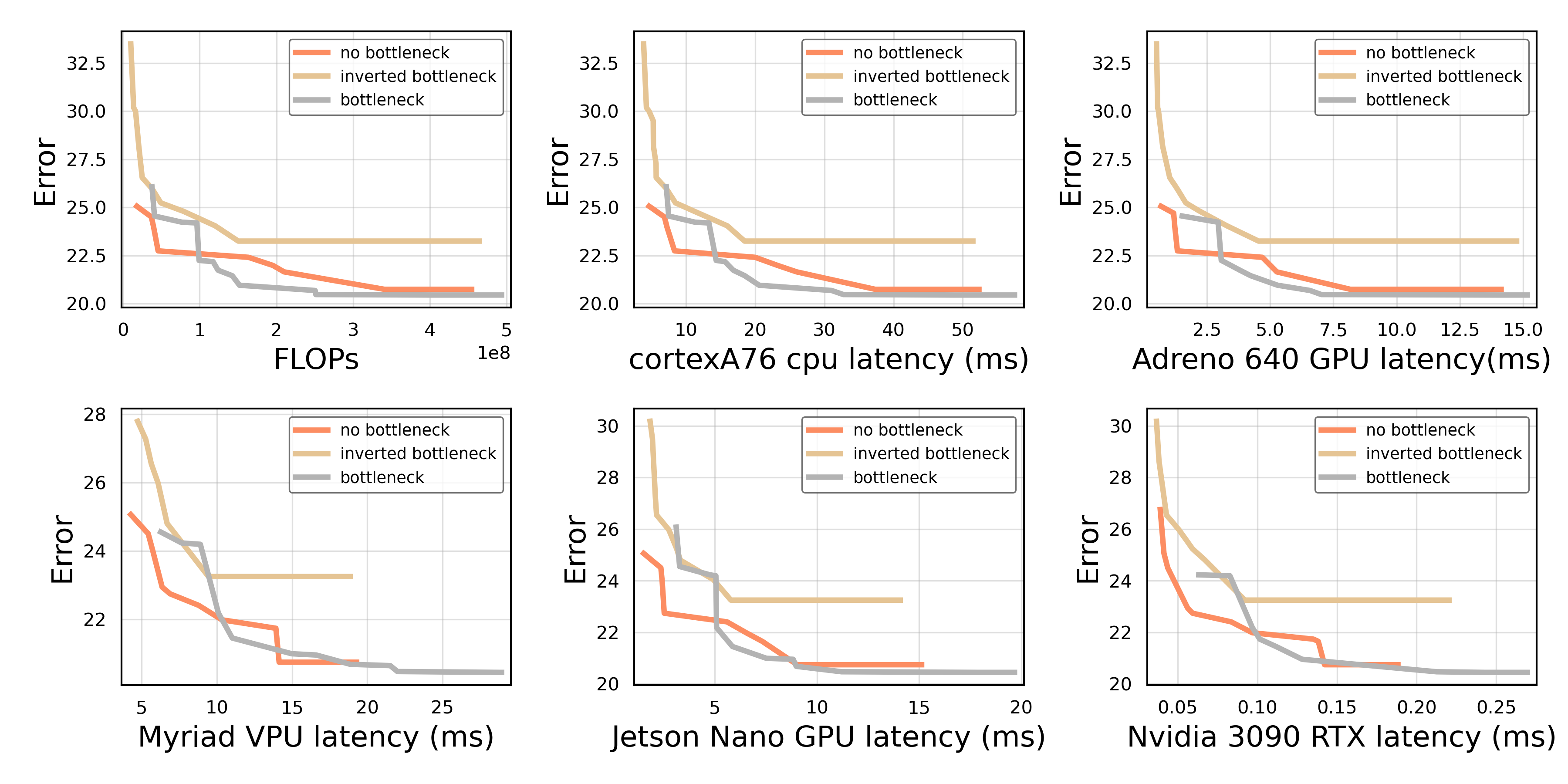}
    \caption{The bottleneck structure has little influence on the latency across all tested hardware platforms. The accuracy of inverted bottlenecks significantly lacks behind building blocks with standard or no bottleneck.}
\label{fig:bottleneck_results}
\end{figure}

The results, as shown in Figure~\ref{fig:bottleneck_results}, indicate that using inverted bottlenecks degrades performance and fails to deliver the promised expected inference speedup. In general, across all hardware platforms, the blocks without a bottleneck and with a regular bottleneck show very similar performance. It can be observed that the VPU and server-grade GPU slightly favour the building block without bottleneck as it executes faster compared to bottleneck blocks. These are the same hardware platforms where the standard convolution outperformed the depthwise separable convolution, indicating that pointwise convolutions underperform on these platforms. 

\subsection{Squeeze-and-Excitation unit}
A Squeeze-and-Excitation (SE) \cite{hu2018squeeze} block is an architectural unit for convolutional neural networks that performs dynamic channel-wise feature recalibration in order to improve the representation power of a convolutional building block.
Given input $\mathbf{X}$ the SE unit first squeezes the spatial information using global average pooling:
\begin{equation} \label{eqn:gap}
    z_c = \frac{1}{H \times W} \sum_{i=1}^{H}\sum_{j=1}^{W} x_c(i, j),
\end{equation}
This step  is followed by an excitation step that recalibrates the channels:
\begin{equation}
    \mathbf{s} = \sigma(W_2(\text{ReLU}(W_1(\mathbf{z})))).
\end{equation}
Where $W_1$ and $W_2$ are two learned linear transformations and $\sigma$ refers to the sigmoid activation function.
The final output $\mathbf{Y}$ of the SE unit is obtained by scaling the original inputs $\mathbf{X}$: $\mathbf{Y} = \mathbf{X}\cdot\mathbf{s}$, where $\cdot$ refers to the channel-wise multiplication.

MnasNet \cite{tan2019mnasnet} brought the SE unit to mobile vision networks, as they were shown to improve model performance compared to previous state-of-the-art models with similar latency. More recently, however, Squeeze-and-Excitation modules have been removed from mobile vision architectures when they are deployed on embedded ML accelerators. EfficientNet-lite \cite{effcientnetlite} and MobileNetEdgeTPUv2 \cite{akin2022searching}, for example, remove the SE unit as they claim it is not well supported for mobile accelerators such as the Google EdgeTPU.

\begin{figure}[h]
    \centering
    \includegraphics[width=\textwidth]{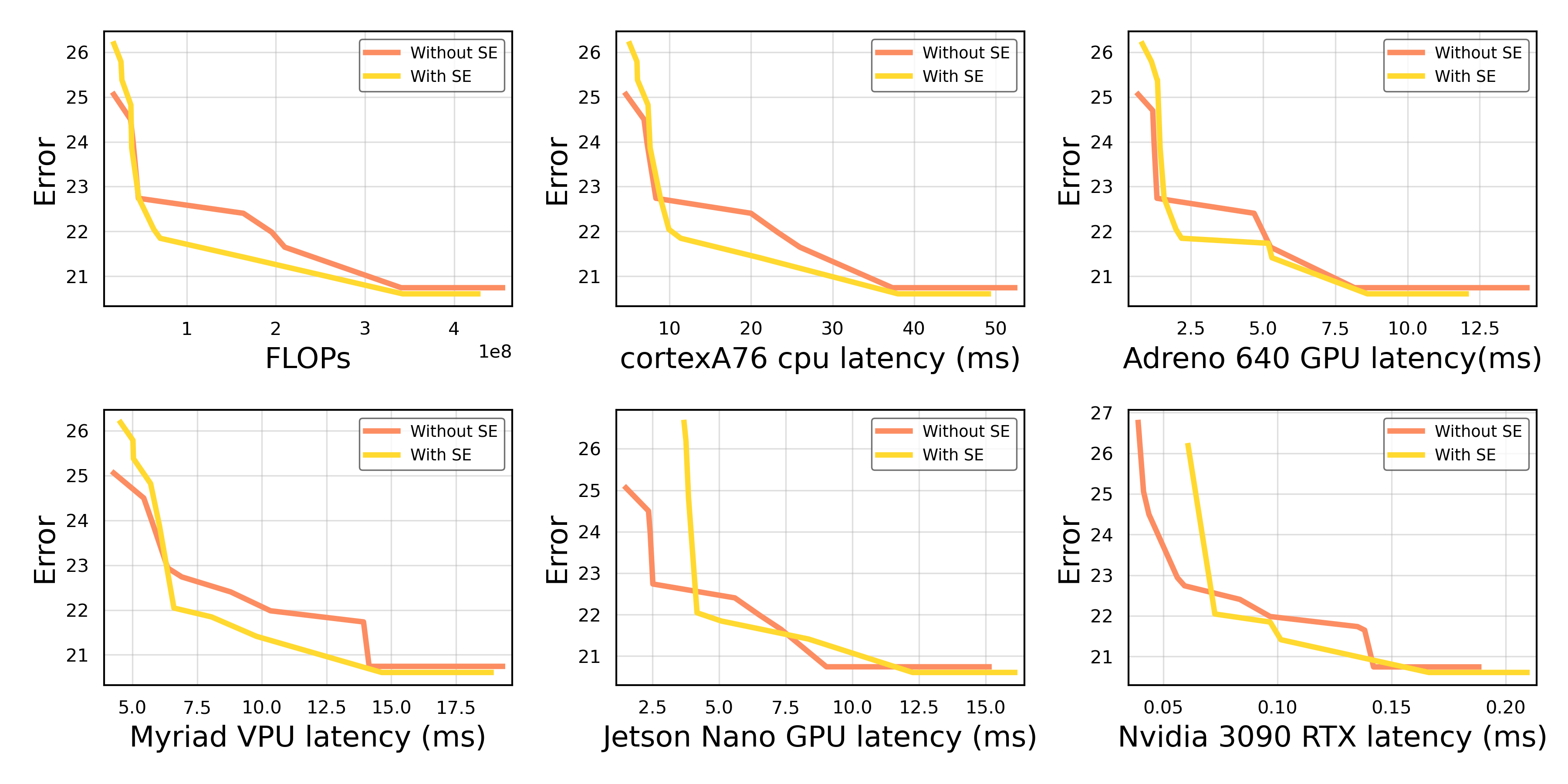}
    \caption{Adding a Squeeze-and-Excitation block improves performance on most hardware platforms but all. The Jetson Nano GPU for example executes them significantly slower, making them not always the optimal choice.}
\label{fig:SE}
\end{figure}

We evaluated the SE block by comparing design spaces with the depthwise separable building block, with and without an SE unit extension.
Figure~\ref{fig:SE} shows that the SE block is indeed a favorable addition for certain mobile platforms as it is very FLOPS-efficient. This also translates to its being more optimal for mobile CPU and GPU. However, for the embedded Jetson Nano GPU it is no longer a clear benefit to include the SE unit as it fails to exploit the optimizations for convolutional kernels and adds a significant delay.
Comparing models with and without the SE unit, and with equal execution time on a mobile CPU, we found that those same models without the SE unit have a running time that is up to a factor $1.9\times$ longer on the embedded Jetson Nano GPU.

\section{Conclusion}
Developing a new model for a targeted hardware platform and with given design constraints requires careful consideration of the building block. An evaluation based on FLOPs often often leads to misleading conclusions w.r.t. the relative benefits of the components in the standard MBConv block on specific hardware platforms. We developed a methodology that can be used to select building blocks and constrain the design spaces for network design (e.g. as an input to NAS) in a hardware-aware manner, while maintaining the freedom to trade off between task performance and execution efficiency. 

We used our approach to evaluate the hardware-efficiency of different convolutional building blocks on various hardware platforms.  Our results show that execution of building blocks with components that are theoretical efficient such as the SE unit take a factor $1.9 \times$ longer to execute than their non-optimized counterparts due to better hardware utilization on platforms with specific embedded ML accelerators such as the intel NCS2 or the Nvidia Jetson Nano.

In essence, our gained insights highlight the importance of benchmarking the building blocks used in mobile vision neural networks, which has been overlooked in the past. 
We believe that our methodology will be key to the develop hardware-aware neural networks for deployment on edge devices.

\section*{Acknowledgments}
This research received funding through the Research Foundation Flanders (FWO-Vlaanderen) under Grant G006718N and 1S47820N.

\bibliographystyle{ACM-Reference-Format}
\bibliography{references}

\end{document}